# Surrogate Model Assisted Cooperative Coevolution for Large Scale Optimization

Zhigang Ren [1], Bei Pang [1], Yongsheng Liang [1], An Chen[1], Yipeng Zhang[1]

[1]*Autocontrol Institute, Xi'an Jiaotong University, Xi'an 710049, China*

**Corresponding author**: Zhigang Ren, Autocontrol Institute, Xi'an Jiaotong University, Xi'an 710049, China

**E-mail**: renzg@mail.xjtu.edu.cn

**Acknowledgements**

This work was supported in part by the National Natural Science Foundation of China under Grant 61105126 and in part by the Postdoctoral Science Foundation of China under Grants 2014M560784 and 2016T90922.

**Abstract** It has been shown that cooperative coevolution (CC) can effectively deal with large scale optimization problems (LSOPs) through a divide-and-conquer strategy. However, its performance is severely restricted by the current context-vector-based sub-solution evaluation method since this method needs to access the original high dimensional simulation model when evaluating each sub-solution and thus requires many computation resources. To alleviate this issue, this study proposes a novel surrogate model assisted cooperative coevolution (SACC) framework. SACC constructs a surrogate model for each sub-problem obtained via decomposition and employs it to evaluate corresponding sub-solutions. The original simulation model is only adopted to reevaluate some good sub-solutions selected by surrogate models, and these real evaluated sub-solutions will be in turn employed to update surrogate models. By this means, the computation cost could be greatly reduced without significantly sacrificing evaluation quality. To show the efficiency of SACC, this study uses radial basis function (RBF) and success-history based adaptive differential evolution (SHADE) as surrogate model and optimizer, respectively. RBF and SHADE have been proved to be effective on small and medium scale problems. This study first scales them up to LSOPs of 1000 dimensions under the SACC framework, where they are tailored to a certain extent for adapting to the characteristics of LSOP and SACC. Empirical studies on IEEE CEC 2010 benchmark functions demonstrate that SACC significantly enhances the evaluation efficiency on sub-solutions, and even with much fewer computation resource, the resultant RBF-SHADE-SACC algorithm is able to find much better solutions than traditional CC algorithms.

# 1 Introduction

Nowadays, large scale optimization problems (LSOPs) are becoming more and more popular in scientific research and engineering applications with the rapid development of big data techniques [1, 2]. Since this kind of problems generally possesses black-box characteristics, the gradient-free evolutionary algorithms (EAs) are often employed to tackle them. However, the performance of conventional EAs rapidly deteriorates as the problem dimension increases. This is the so-called 'curse of dimensionality' [3, 4], the main reason for which consists in that the solution space of a problem exponentially grows with the increase of its dimension and conventional EAs cannot adequately explore the solution space of a LSOP within acceptable computation time.

Taking the idea of 'divide-and-conquer', cooperative coevolution (CC) provides a natural way for solving LSOPs [5]. It first decomposes an original LSOP into several smaller and simpler sub-problems, and then solves the LSOP by cooperatively optimizing all the sub-problems with a conventional EA. It is understandable that decomposition plays a fundamental role in CC. A right decomposition can greatly reduce the optimization difficulty of a LSOP without changing its theoretical optimum. Therefore, in recent years, most research efforts on CC were put into designing various kinds of decomposition methods, and by now several efficient decomposition algorithms have been developed [6].

By contrast, another important algorithmic operation in CC, i.e., the evaluation of the solutions to sub-problems, which also has an important influence on the efficiency of CC, is neglected. It is known that CC mainly focuses on black-box LSOPs which has no explicit objective functions and generally evaluates their solutions by simulation. This means that all the sub-problems obtained through decomposition do not own separate or explicit objective functions. To evaluate sub-solutions, now all the CC algorithms adopt a context-vector-based method [4]. This method takes a complete solution of the original LSOP as context vector. For a sub-solution to be evaluated, the method first inserts it into the corresponding positions in the context vector, and then achieves evaluation by indirectly evaluating the modified context vector with the simulation model of the original LSOP. This method seems straightforward and reasonable. It is also in this way that different sub-problems cooperate with each other. Nevertheless, it also brings some issues. First, a limited number of solution simulations are generally allowed for a practical LSOP since even a single simulation is very time-consuming, then the number of simulations assigned to each sub-problem will further reduce. With so few computation resource, it is challenging for the optimizer in CC to generate high quality sub-solutions. As a result, the quality of the final solution to the original LSOP can hardly be guaranteed. Second, for an incorrectly decomposed LSOP, it is theoretically impossible to find its global optimum via a single context vector. Multi-context-vector-based evaluation method is likely to remedy this defect [7-9], but it may further raise solution simulation requirements. In this case, it is very significant to develop an efficient sub-solution evaluation method which depends less on the original simulation model.

As explained above, the sub-problems obtained through decomposition can be regarded as small or medium scale computationally expensive black-box optimization problems. To deal with this kind of problems, surrogate model assisted EAs (SAEAs) were developed [10-12]. Their key idea is to construct a calculable surrogate model for the computationally expensive objective function and employ the surrogate model to evaluate solutions. Only some promising solutions filtered by the surrogate model need to access the real objective function. By this means, the number of real evaluations can be greatly reduced. So far, several types of surrogate models have been proposed and integrated with a variety of EAs [11-12]. However, these SAEAs only take effect on low and medium dimensional problems. The main reason consists in that a high dimensional problem often requires too many real evaluated solutions to build an accurate enough surrogate model.

This study first introduces the surrogate model technique into the CC framework and develops a novel surrogate model assisted CC (SACC) framework. By constructing and maintaining a surrogate model for each sub-problem, SACC is expected to solve two main issues concerned in the traditional CC and SAEA. From the point of CC, surrogate model improves the sub-solution evaluation efficiency and reduces the requirement on the solution simulation times. From the point of SAEA, CC reduces the difficulty of training surrogate model by decreasing the problem dimension via decomposition and scales up surrogate model and EA to LSOPs. To sum up, surrogate model and CC complement each other well within the SACC framework. To show the efficiency of SACC, this study implements a concrete SACC algorithm which takes radial basis function (RBF) [13] and success-history based adaptive differential evolution (SHADE) [14] as surrogate model and optimizer, respectively, and names the resultant algorithm RBF-SHADE-SACC. RBF and SHADE have been proved to be effective for small and medium scale problems, but have never been employed to solve LSOPs. This study first scales them up to LSOPs of up to 1000 dimensions with the help of SACC and also tailors them to make them adapt to the characteristics of LSOP and SACC.

The rest of this paper is organized as follows. Section 2 presents the related work on CC and surrogate model. Section 3 describes SACC in detail, including the framework of SACC, the tailored RBF and SHADE, and the final RBF-SHADE-SACC algorithm. Section 4 reports experimental studies. Finally, conclusions are drawn in section 5.

# 2 Related work

This section first reviews existing CC algorithms, then briefly introduces the commonly-used surrogate models in SAEAs.

## 2.1 Cooperative coevolution

CC can effectively tackle LSOPs by cooperatively optimizing the lower dimensional sub-problems obtained through decomposition. It is understandable that decomposition plays a pivotal role in CC. So far, many decomposition algorithms have been developed [6]. They can be generally divided into three categories, including static decomposition, random decomposition, and learning-based decomposition. Static decomposition is the simplest type of decomposition method in which decision variables are divided into a certain number of fixed groups. Potter and De Jong [5] proposed the first CC algorithm which partitions an *n*-dimensional problem into *n* independent 1-dimensional sub-problems. They also developed a splitting-into-half decomposition strategy which decomposes an *n*-dimensional problem into two fixed $n/2$ dimensional sub-problems [15]. More generally, Van den Bergh and Engelbrecht [4] suggested grouping an *n*-dimensional problem into *k* *s*-dimensional sub-problems for some $ks=n$ and $s \ll n$. These static decomposition strategies perform well on separable problems, but usually show poor performance on nonseparable problems as they take no account of variable interactions.

To remedy this defect, some random decomposition methods were developed. Yang *et al.* [16] proposed the first random decomposition method and named it random grouping. This method randomly allocates all the decision variables into *k* *s*-dimensional sub-problems in every coevolution cycle instead of using a static grouping. To tackle the issue that it is difficult to specify a value for *s*, Yang *et al.* [17] further developed a multilevel CC algorithm which selects a value for *s* from a pool for each new coevolution cycle with a higher probability if this value helps the CC algorithm achieve greater performance improvement in the last cycle when it is adopted. Omidvar *et al.* [18] indicated that, with random grouping, the probability of grouping all the interacting variables into one sub-component dramatically reduces as the number of interacting variables increases, and suggested increasing the frequency of random grouping by reducing the iteration times within a cycle.

In recent years, some learning-based decomposition methods were developed. They focus on making near optimal decomposition by explicitly detecting the interdependencies among variables. Delta grouping [19] can be regarded as an early representative of this kind of decomposition method. It calculates the variations of variables in two consecutive cycles and divides the variables with similar variations into the same sub-component. It was shown that delta grouping can outperform random grouping on a variety of LSOPs, but often loses its efficiency on the problems having more than one group of interacting variables [19, 20]. To overcome this limitation, Omidvar *et al.* [20] proposed a differential grouping (DG) method. DG regards two variables as separable if one variable does not influence the change of the fitness value (FV) caused by the change of the other variable. However, the original DG ignores indirect interdependencies among variables and may group some interacting variables into different sub-components. Aiming at this shortcoming, global DG [21] was proposed. It explicitly detects the interdependency between each pair of variables. Consequently, the decomposition accuracy is greatly improved at the cost of consuming much more computation resources. Recently, Omidvar *et al.* [22] developed a new version of DG which reduces the computation resource requirement by reusing some samples. Ren *et al.* [23] further improved the decomposition efficiency of DG by detecting the interdependencies from the point of vectors.

After decomposition, CC needs a specific algorithm to optimize the obtained sub-problems. By now, almost all kinds of EAs, such as genetic algorithm (GA) [5], particle swarm optimization (PSO) [4, 24], and differential evolution (DE) [7, 8], have been employed as optimizer in CC. No matter which kind of algorithm is used, the solutions of each sub-problem need to be evaluated. As indicated in the introduction section, each sub-problem does not have a separate or explicit objective function, and its solutions also cannot be directly evaluated by the simulation model of the original LSOP since they only reflect part of dimensions of the total solution space. Now all the existing CC algorithms adopt a context-vector-based method to evaluate sub-solutions [4]. This method takes a complete solution of the original LSOP, which is generally the best solution obtained so far, as the context vector. For a sub-solution to be evaluated, this method first inserts it into the corresponding positions in the context vector, and then achieves the evaluation by indirectly evaluating the modified context vector with the original simulation model. It has been verified that the context-vector-based method really takes effect [4, 6], but it requires many computation resources since it needs to invoke the original simulation model when evaluating each sub-solution.

On the other hand, this method is generally performed based on a single context vector, which may hinder optimizer from finding the optimal solution for an incorrectly decomposed LSOP [7]. To alleviate this issue, some multi-context-vector-based evaluation methods were developed. Wu *et al.* [7] and Tang *et al.* [8] proposed two multi-context mechanisms of similar idea. Both mechanisms maintain a context set containing a certain number of complete solutions. When evaluating a sub-solution, they randomly select a solution from the context set as the context vector. After each coevolution cycle, a crossover operation may be performed on the worst context vector to improve its performance. Peng *et al.* [9] suggested solving each sub-problem in CC with a multi-population scheme and constructing multiple context vectors with the optima achieved by the multiple populations. For a sub-solution to be evaluated, all the context vectors are used and the sub-solution is finally evaluated according to the obtained best fitness. There is no doubt that the multi-context-vector-based evaluation methods have some theoretical advantages over the single-context mechanism. Nevertheless, if straightforwardly performed, they are much likely to consume much more computation resources. Therefore, it becomes more and more important to develop more efficient sub-solution evaluation methods.

## 2.2 Surrogate model

It is known that most EAs require a large number of fitness evaluations (FEs) before locating the global optimal or near-optimal solution. However, some computationally expensive optimization problems cannot support enough number of FEs. To deal with this kind of problems, SAEAs were developed [10-11]. SAEA builds a surrogate model for the computationally expensive objective function and employs the model to evaluate the candidate solutions. Only some promising solutions filtered by the surrogate model need to access the real objective function. By this means, the requirement on the real FEs can be greatly reduced. SAEAs received increasing attentions in recent years, and several types of surrogate models, including polynomial regression (PR), support vector regression (SVR), Gaussian process (GP) regression, and RBF, were developed [12, 13, 25]. Among these models, PR is easy to train but is generally of low estimation accuracy,

SVR is able to relieve the 'curse of dimensionality' but has difficulty in tackling large scale samples, GP can fit complex response surface well but asks long training time and shows dramatic performance deterioration as the problem dimension increases. As for RBF, it is easy to train and is relatively robust to the change of problem dimension [12]. Profiting from these excellent performance, now RBF is most widely employed by SAEAs.

No matter which type of surrogate model is adopted, it is essentially impossible to build a surrogate model that can correctly evaluate all the solutions generated during the whole optimization process due to the high dimensionality, ill distribution, and limited number of training samples. To weaken the negative influence of the evaluation error, it is necessary to evaluate part of solutions with the real objective function. This brings the so-called model management problem, i.e., determining which solutions should be evaluated by the real objective function. Now two classes of model management strategies have been developed. One is the generation-based strategy which employs the real objective function for fitness evaluation at some specified generations [26], and the other is the individual-based strategy which employs the real objective function to evaluate some selected individuals at every generation [27]. These generations and individuals are often specified according to a static rule, a random rule, or even an adaptive rule. It is revealed that the individual-based model management may be more suited for steady state evolution or generational evolution implemented on a single machine. By contrast, the generation-based model management is better for parallel implementation on heterogeneous machines having different speeds [11].

According to the solution region covered by surrogate model, existing SAEAs can be divided into three categories, including global-surrogate assisted EAs [25], local-surrogate assisted EAs [28], and ensemble-surrogate assisted EAs [13]. The global-surrogate model tries to model the whole solution space and generally has strong exploration ability. Its main drawback lies in the difficulty in ensuring the estimation accuracy. The local-surrogate model aims at the current search region of EA. Compared with the global-surrogate model, it is more likely to produce more accurate fitness estimations, but can hardly help EA escape from local optima. By contrast, the ensemble-surrogate model take the advantages of both the global and the local surrogate models by integrating them together and was shown to be able to outperform a single kind of model in most cases.

It has been proved that SAEAs really take effect on computationally expensive optimization problems. However, they were mainly verified on small scale problems of dimensions lower than 30 [10, 13]. In recent years, some researchers tried to scale up surrogate models to medium scale problems. Liu *et al.* [25] proposed a GP assisted EA which achieves good performance on problems of 50 dimensions with the help of a dimension reduction technique. To the best of our knowledge, the highest dimension tackled by SAEAs is 100. The corresponding algorithm is surrogate-assisted cooperative swarm optimization [29] which cooperatively employs a surrogate-assisted PSO and a surrogate-assisted social learning based PSO to search for the global optimum. However, with respect to large scale problems of dimensions up to 1000, all the existing surrogate models will lose their efficiency if they are applied in a straightforward way. This paper makes the first attempt to scale up surrogate model to LSOPs of 1000 dimensions by introducing it into CC. Within the CC framework, it is much possible to construct sufficiently accurate surrogate models for lower dimensional sub-problems obtained through decomposition, and these surrogate models can be expected to significantly improve the evaluation efficiency of a vast number of sub-solutions generated during the CC process.

## 3 Description of SACC

Many real world LSOPs are difficult to tackle but possess an appealing feature, i.e., separability, where partially additive separability is the most common type and is most extensively studied in the CC research field [20-23]. The definition of additive separability can be described as follows.

**Definition 1** A function is partially additively separable if it has the following general form:

$$f(\boldsymbol{x}) = \sum_{g=1}^{k} f_g(\boldsymbol{x}_g),  \tag{1}$$

where $\boldsymbol{x} = (\boldsymbol{x}_1, \boldsymbol{x}_2, ..., \boldsymbol{x}_k)$ is a global decision vector of $n$ dimensions, $\boldsymbol{x}_g (g=1,2,...,k)$ are mutually exclusive decision vectors of $f_g(\cdot)$, and $k$ is the number of independent sub-components [20].

CC solves a high dimensional problem $f(\boldsymbol{x})$ by first identifying all the mutually exclusive sub-components $\boldsymbol{x}_g (g=1,2,...,k)$ and then cooperatively optimizing the resultant lower dimensional sub-problems $f_g(\boldsymbol{x}_g)$. Fig. 1 shows the general framework of the traditional CC, where it first initializes the parameters of a specified optimizer, the best overall solution $\boldsymbol{x}^*$, and the population $P_g$ for each sub-problem after the decomposition operation, then it enters an iterative process. During each iteration, a sub-problem $g$ is first selected generally in a round-robin fashion, then a child population $U_g$ is generated for the selected $g$th sub-problem based on its current population $P_g$. After evaluation, some good solutions in $U_g$ are finally employed to update $P_g$ and $\boldsymbol{x}^*$. This process of selection, optimization, and update is iterated until all the available computation resources are exhausted.

It is notable that the expression of each $f_g(\cdot)$ is unknown as $f(\cdot)$ is a black-box optimization problem in the context of CC. To evaluate the solutions of each sub-problem, CC maintains a context vector which is generally set as the best overall solution $\boldsymbol{x}^*$ [4]. More concretely, it evaluates a sub-solution $\boldsymbol{x}_g$ by means of $f(\boldsymbol{x}^* | \boldsymbol{x}_g)$, where $\boldsymbol{x}^* | \boldsymbol{x}_g$ denotes the complete solution that replaces the corresponding sub-component of $\boldsymbol{x}^*$ with $\boldsymbol{x}_g$. This context-vector-based evaluation method needs to access the original high dimensional simulation model when evaluating each low dimensional sub-solution and thus requires many computation resources. On the other side, the available computation resources are very limited for real word LSOPs, then traditional CC can hardly obtain high quality solutions in this case. Besides, when updating the current sub-population, some optimizers, such as the commonly employed DE, need to compare the individuals therein with the ones in the corresponding child sub-population. In general, current individuals have been evaluated at past iterations, when the context vector, i.e. $x^*$, may be different from the current one $x^{*'}$. In this case, the FV of a current individual $f(x^* | x_g^c)$ and the one of a new individual $f(x^{*'} | x_g^n)$ are incommensurable. A straightforward way to tackle this issue is to reevaluate current individuals if the context vector is really updated, but this will consume extra computation resources. Aiming at this issue, traditional CC persistently optimizes a selected sub-problem for a certain number of iterations, during which the context sub-vectors provided by the other sub-problems remain unchanged [6, 16]. However, this approach still needs to reevaluate the individuals of a sub-problem when it is just selected to optimize. It also limits the interaction frequency among different sub-problems, which is adverse to improving the performance of CC [18].

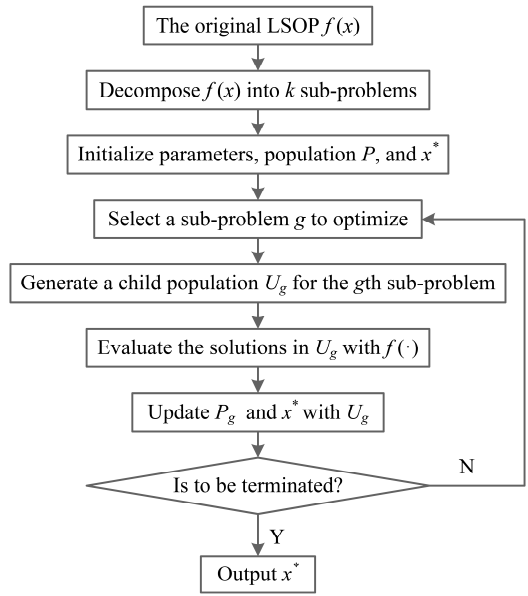

**Fig. 1** The general framework of the traditional CC

## 3.1 The framework of SACC

According to the above explanation, how to efficiently evaluate sub-solutions becomes a key issue in the traditional CC. This issue has the following two characteristics: 1) Compared with the original LSOP, the sub-problems obtained via decomposition are usually small and medium scale computationally expensive problems; 2) The goal of evaluation is not to get the accurate FV of each sub-solution, but to identify which one is better from a certain number of candidates. It has been shown that surrogate models fit well to this kind of problems, but they have never been applied to LSOPs. As the first attempt, SACC scales up surrogate models to LSOPs with the help of CC. To achieve this, SACC maintains a surrogate model for each sub-problem and mainly depends on the surrogate model to evaluate sub-solutions. The original simulation model is only adopted to reevaluate fewer good sub-solutions selected by the surrogate model, and these sub-solutions will be in turn employed to update the surrogate model. As a result, the number of real FEs could be greatly reduced without significantly sacrificing evaluation quality.

Fig. 2 presents the framework of SACC, from which it can be seen that SACC shares same operation modules with the traditional CC except the surrogate model construction module and the sub-solution evaluation module. Besides the population $P_g$, SACC also maintains an external archive $D_g$ for each sub-problem $g$ to record a certain number of real evaluated sub-solutions. These sub-solutions are continually updated and further employed to update the surrogate model such that the estimation accuracy of the model could be increased as high as possible. SACC provides a general framework that makes surrogate model and CC complement each other. To implement a concrete SACC algorithm, it is necessary to specify the type of surrogate model, the model management strategy, and the optimizer for each sub-problem. We will describe them in the following sub-sections.

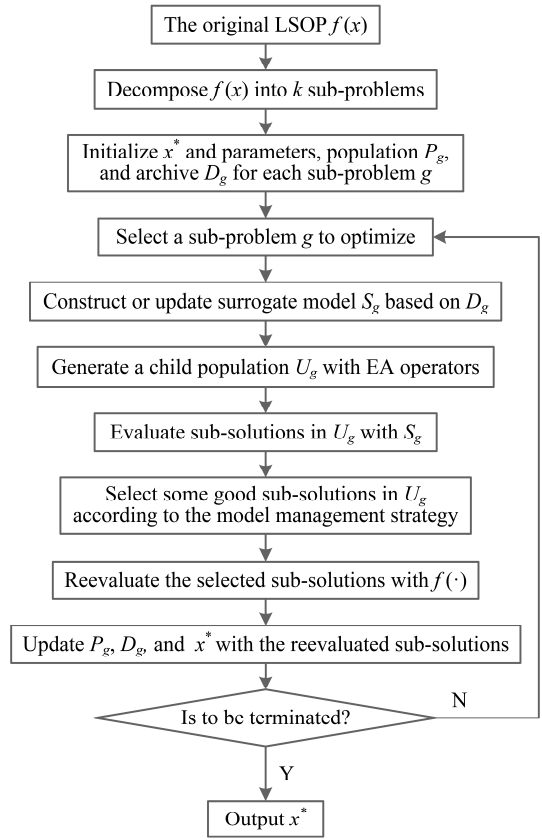

**Fig. 2** The framework of SACC

## 3.2 RBF model for SACC

As reviewed in section 2.2, several types of surrogate models have been developed by now, among which RBF is easier to train, more robust to different problem dimensions, and is also more widely applied in the traditional SAEAs [13, 30-32]. Considering all these advantages, this study employs the RBF model described in [30] as the surrogate model for SACC. RBF is essentially an instance-based learning method. Given $d$ training samples $t_g^1, t_g^2, \cdots, t_g^d \in \mathbb{R}^s$ for a sub-problem $g$ of $s$ dimensions, the evaluation value provided by RBF for a new sub-solution $x_g \in \mathbb{R}^s$ can be represented as

$$\overline{e}(x_g) = \sum_{i=1}^{d} \omega_i \phi(\| x_g - t_g^i \|) + \boldsymbol{\beta}^T x_g + \alpha, \tag{2}$$

where $\| \boldsymbol{x}_g - \boldsymbol{t}_g^i \|$ denotes the Euclidean distance between the two sub-solutions, $\phi(\cdot)$ denotes the basis function, and $(\omega_1, \omega_2, \cdots, \omega_d)^T = \boldsymbol{\omega} \in \mathbb{R}^d$, $\boldsymbol{\beta} \in \mathbb{R}^s$, and $\alpha \in \mathbb{R}$ are corresponding parameters. There are several different choices for the basic function $\phi(\cdot)$, such as cubic basis function, Gaussian basis function, and multi-quadric function [31]. This study adopts cubic basis function, i.e., $\phi(r) = r^3$, since it was shown to be more suitable than other functions for SAEAs. As for $\boldsymbol{\beta}^T \boldsymbol{x}_g + \alpha$, it is a polynomial tail appended to the standard RBF.

With the given training samples, the parameters $\boldsymbol{\omega} \in \mathbb{R}^d$, $\boldsymbol{\beta} \in \mathbb{R}^s$, and $\alpha \in \mathbb{R}$ can be obtained by solving the following linear system of equations:

$$\begin{pmatrix} \Phi & Q \\ Q^T & \mathbf{0} \end{pmatrix} \begin{pmatrix} \boldsymbol{\omega} \\ \boldsymbol{\gamma} \end{pmatrix} = \begin{pmatrix} \boldsymbol{e} \\ \mathbf{0} \end{pmatrix}, \tag{3}$$

where $\Phi$ is a $d \times d$ matrix with $\Phi_{ij} = \phi(\| \boldsymbol{t}_g^i - \boldsymbol{t}_g^j \|)$, $\boldsymbol{\gamma} = (\boldsymbol{\beta}^T, \alpha)^T$, $\boldsymbol{e} = (e(\boldsymbol{t}_g^1), e(\boldsymbol{t}_g^2), \cdots, e(\boldsymbol{t}_g^d))^T$ with each $e(\boldsymbol{t}_g^i)$ being the real evaluation value of the *i*th training sample, and

$$Q^T = \begin{pmatrix} \boldsymbol{t}_g^1 & \boldsymbol{t}_g^2 & \cdots & \boldsymbol{t}_g^d \\ 1 & 1 & \cdots & 1 \end{pmatrix}. \tag{4}$$

It is proved that the square matrix in the left hand of (3) is invertible if and only if $\text{rank}(Q) = s + 1$ [30]. In this case, the linear system of equations has a unique solution, then (2) can be used as the approximate model to evaluate a new sub-solution $\boldsymbol{x}_g \in \mathbb{R}^s$.

It is notable that different from the traditional SAEAs which usually employ surrogate model to estimate the FV of a new solution, SACC adopts RBF to predict the fitness improvement made by a new sub-solution $\boldsymbol{x}_g$ to the best overall solution $\boldsymbol{x}^*$. The reasons are twofold: 1) Each sub-problem does not have an explicit objective function; 2) The evaluation values $f(\boldsymbol{x}^* | \boldsymbol{x}_g)$ of sharply different solutions $\boldsymbol{x}_g$ to the same sub-problem generally show small differences since each $f(\boldsymbol{x}^* | \boldsymbol{x}_g)$ adds the FVs of the solutions to all the sub-problems together, which makes against constructing an accurate RBF model. For a minimization problem, the fitness improvement of a sub-solution $\boldsymbol{x}_g$ is defined as $e(\boldsymbol{x}_g) = f(\boldsymbol{x}^*) - f(\boldsymbol{x}^* | \boldsymbol{x}_g)$. According to the property of additive separability defined by (1), it eliminates the influence of other sub-problems and enlarges the relative differences among the evaluation values of different solutions to the same sub-problem. Obviously, a sub-solution of larger fitness improvement is considered better.

As indicated above, for an *s*-dimensional sub-problem, $\text{rank}(Q) = s + 1$ must hold to generate a unique group of parameters for its RBF model. This requires the number of training samples $d \geq s + 1$. How to set *d* based on this basic condition affects much on the performance of SACC since it usually involves many sub-problems of different dimensions. For a given sub-problem, a larger *d* generally leads to a more accurate RBF model, but it also asks for more computation resources and may result in overfitting. To weaken the sensitivity of RBF to *d*, SACC does not build a fixed RBF model for each sub-problem, but first constructs an initial model and then continually updates it with new samples. To achieve this, SACC maintains an external archive $D_g$ for each sub-problem $g$ to store the newest $d$ sub-solutions evaluated by the original simulation model and the corresponding fitness improvement values. $D_g$ is generally initialized with $d$ randomly generated sub-solutions and is continually updated in an iteration-wise way. Once $D_g$ is updated, the RBF model is retrained accordingly. By this means, SACC can get a coarse global RBF model for each sub-problem at its initial search stage which could filter out some local optima, and the model gradually turns to a more and more accurate local model as new samples are employed. It is notable that the sub-solutions in an archive are generally introduced in different iterations, during which the context vectors, i.e., $x^*$, may change. Without fine intervention, this would make the fitness improvements of these sub-solutions incommensurable. To avoid this, once a better context vector $x^{*'}$ is found, SACC updates the fitness improvement values of all the sub-solutions in current $D_g$ according to the following way:

$$\text{if } \Delta f = f(x^*) - f(x^{*'}) > 0, \text{then } e(x_g) \leftarrow e(x_g) - \Delta f, \text{for } \forall x_g \in D_g. \tag{5}$$

## 3.3 SHADE for SACC

SHADE is an excellent DE variant developed in recent years [14]. It inherits the efficient 'current-to-$p$best/1' mutation operator from the classic JADE algorithm [33], but further improves JADE with a novel parameter adaptation mechanism. Instead of employing a single pair of parameter means to generate the new mutation factor and crossover rate as JADE, SHADE maintains a diverse set of means for each parameter to guide its adaptation. In this way, the negative impact of some poor parameter values can be reduced. SHADE and its variants achieved great success in solving small and medium scale optimization problems [34], but have seldom been adopted to solve LSOPs. This study scales up SHADE for LSOPs under the framework of SACC by employing it as optimizer for the lower dimensional sub-problems obtained through decomposition. SACC not only keeps the key features of SHADE, but also tailors it to make it adapt to the characteristics of CC and surrogate model. For the convenience of description, we name the tailored SHADE as $t$SHADE.

For each sub-problem $g$, $t$SHADE maintains a population $P_g$ which contains $p$ real evaluated individuals. As the original SHADE, $t$SHADE produces a trial vector $u_g^i$ ($i=1,2,\cdots,p$) for each individual $x_g^i$ in $P_g$ at each generation according to the following 'current-to-$p$best/1' mutation operator and binomial crossover operator:

$$v_g^i = x_g^i + F_i(x_g^{pbest} - x_g^i) + F_i(x_g^{r1} - \tilde{x}_g^{r2}), \tag{6}$$

$$u_g^{i,j} = \begin{cases} v_g^{i,j}, & \text{if } rand(0,1) \le CR_i \text{ or } j = j_{rand} \\ x_g^{i,j}, & \text{otherwise} \end{cases}, \tag{7}$$

where $v_g^i$ denotes the mutation vector of $x_g^i$, $u_g^{i,j}$ denotes the $j$th element of $u_g^i$, and all the other symbols have the same meanings as the corresponding ones in the original SHADE. It is notable that the mutation factor $F_i$ in (6) and the crossover rate $CR_i$ in (7) are individual-dependent. They are generated based on a historical memory $M_g$ which contains a certain number of pairs of their means. Besides, $t$SHADE also maintains an external archive $A_g$ which preserves some inferior solutions and together with $P_g$ provides candidates for $\tilde{x}_g^{r2}$. The update of $M_g$ and $A_g$ depends heavily on the relative quality of each pair of $x_g^i$ and $u_g^i$. At each generation, an entry in $M_g$ is updated in a round-robin fashion by the respective means of all the effective $F_i$ and $CR_i$ ($i=1,2,\cdots,p$) that help $u_g^i$ outperform the corresponding $x_g^i$. On the other side, all the failing $x_g^i$ are employed to fill or update $A_g$.

At this moment, the key issue is to effectively judge whether a $u_g^i$ is better than the corresponding $x_g^i$. For the original SHADE, this can be easily achieved by directly comparing the FVs of the two solutions. Nevertheless, this approach is unacceptable for $t$SHADE since SACC does not allow $t$SHADE to really evaluate each $u_g^i$. On the contrary, if we compare all pairs of $u_g^i$ and $x_g^i$ completely depending on the constructed RBF model, then the comparison result will be affected too much by the evaluation error of RBF, especially at the initial search stage of SACC. $t$SHADE fulfils this task in a two-step manner. First, it compares each pair of $u_g^i$ and $x_g^i$ based on the evaluation results $\bar{e}(u_g^i)$ and $\bar{e}(x_g^i)$ provided by RBF. Second, it picks out $q$ ($q \ll p$) best trial vectors, reevaluates them with the original simulation model, and adjusts the comparison results on the corresponding population individuals according to the new evaluation results. This two-step manner avoids most real evaluations on the one hand, and also ensures the evaluation accuracy on high quality sub-solutions on the other hand.

As for the population update operation, in contrast to the original SHADE which always replaces a population individual with the corresponding successful trial vector, $t$SHADE neglects the correspondence between a population individual and its trial vector, and updates some bad individuals with the $q$ reevaluated trial vectors if the latter achieve larger fitness improvements. By this means, the population of each sub-problem essentially keeps the $p$ best sub-solutions that are really evaluated by the original simulation model, and the risk brought by the evaluation error of RBF can be greatly reduced.

As indicated at the end of section 3.2, the RBF archive $D_g$ of each sub-problem $g$ needs to be updated in time to learn a more and more accurate RBF model. SACC achieves this by replacing the $q$ oldest individuals in $D_g$ with the $q$ new sub-solutions really evaluated at each generation. Besides, it is worth mentioning that once the best overall solution $x^*$ is improved, the evaluation values of all the individuals in $D_g$ and $P_g$ should be updated according to (5) such that these individuals are commensurable. It is also in this way that the reevaluation issue concerned in the traditional CC can be eliminated.

## 3.4 The procedure of RBF-SHADE-SACC

Integrating RBF and $t$SHADE into the SACC framework, we can get the procedure of RBF-SHADE-SACC as presented in Algorithm 1. Steps 2-7 mainly perform initialization operations, where step 5 initializes the parameter memory $M_g$ with the same rule as the one in the original SHADE [14], but initializes the external archive $A_g$ in a different way. The original SHADE initializes its external archive with an empty set and continually fills or updates the archive with population individuals which are worse than the corresponding trial vectors. This asks SHADE to perform different operations according to the real number of solutions in the archive during the whole evolution process, although the archive will be definitely fully filled after several generations. To simplify these operations, Step 5 directly initializes $A_g$ with a specified number of random sub-solutions. This minor change significantly simplifies the implementation of SHADE without affecting its performance. As for the external archive $D_g$ of RBF and the population $P_g$ of $t$SHADE, they require $d$ and $p$ real evaluated sub-solutions for initialization, respectively. To reduce the number of real evaluations, step 6 directly generates $\max(d,p)$ sub-solutions, which are further assigned to $D_g$ and $P_g$ in step 7.

**Algorithm 1**: RBF-SHADE-SACC
1. Generate a decomposition $x \to \{x_1, x_2, \cdots, x_k\}$;
2. Initialize the best overall solution $x^*$ with a randomly generated complete solution;
3. **for** sub-problem $g = 1: k$ **do**
4.     Initialize the parameters of RBF and $t$SHADE, including $d$, $p$, and $q$;
5.     Initialize all the entries in $M_g$ to 0.5 and initialize $A_g$ with $p$ random sub-solutions;
6.     Randomly generate $\max(d,p)$ sub-solutions $x_g$ and evaluate them with $e(x_g)$;
7.     Initialize $D_g$ and $P_g$ with $d$ and $p$ generated sub-solutions, respectively;
8. Select a sub-problem $g$ to optimize according to a specified rule;
9. Build a RBF model for the $g$th sub-problem with $D_g$ according to (2)-(4);
10. **for** each sub-solution $x_g^i \in P_g$ **do**
11.     Generate a pair of control parameters $F_i$ and $CR_i$ based on $M_g$;
12.     Generate a trial vector $u_g^i$ according to the "current-to-$p$best/1/bin" rule presented by (6)-(7);
13.     Evaluate $x_g^i$ and $u_g^i$ with $\bar{e}(x_g^i)$ and $\bar{e}(u_g^i)$ provided by RBF, respectively;
14. Select $q$ best trial vectors from the group of $u_g^i, i=1,2,\cdots,p$ and store them into $Q_g$;
15. Reevaluate each trial vector $u_g^i \in Q_g$ with $e(u_g^i)$ and modify $\bar{e}(u_g^i) \leftarrow e(u_g^i)$;
16. **for** each sub-solution $x_g^i \in P_g$ **do**
17.     **if** $\bar{e}(u_g^i) > \bar{e}(x_g^i)$ **then**
18.         Replace a randomly selected sub-solution in $A_g$ with $x_g^i$;
19.         Record the corresponding control parameters $F_i$ and $CR_i$;
20. Update an entry in $M_g$ based on the recorded successful control parameters;
21. Update the $q$ oldest sub-solutions in $D_g$ with the trial vectors in $Q_g$;
22. **for** each $u_g^i \in Q_g$ **do**
23.     Find out the worst sub-solution $x_g^w$ in $P_g$;
24.     **if** $e(x_g^w) < e(u_g^i)$ **then**
25.         Delete $x_g^w$ from $P_g$ and insert $u_g^i$ into $P_g$;
26. Find out the best sub-solution $x_g^b$ in $P_g$;
27. **if** $e(x_g^b) > 0$ **then**
28.     Update $f(x^*) \leftarrow f(x^* | x_g^b)$, $x^* \leftarrow x^* | x_g^b$;
29.     **for** each sub-solution $x_g^i \in D_g \cup P_g$ **do**
30.         Update $e(x_g^i) \leftarrow e(x_g^i) - e(x_g^b)$;
31. **if** termination condition is not met **then** goto step 8;
32. Output $x^*$, $f(x^*)$.

After the initialization operation, step 8 selects a sub-problem to optimize according to a specified rule. The selection rule essentially determines the computation resource allocation among different sub-problems. Although several different selection rules have been proposed by now [35, 36], this study adopts the basic round-robin rule as the traditional CC to highlight the efficiency of SACC in saving computation resource. Nevertheless, different from the traditional CC which persistently optimizes a selected sub-problem for a certain number of iterations, SACC just allows a selected sub-problem to undergo a single iteration, by which the interaction frequency among different sub-problems can be greatly increased. Steps 10-13 generate a trial vector for each individual in the current sub-population and evaluate each pair of sub-solutions with the RBF model constructed at step 9. Steps 14-20 identify a better sub-solution from each pair of $u_g^i$ and $x_g^i$ according to the two-step manner described in section 3.3, and update $A_g$ and $M_g$ in the light of the identification result. For the concrete update rule of $M_g$, readers can refer to equations (13)-(14) and (17)-(19) in [14]. Steps 21-25 update $D_g$ and $P_g$ with the $q$ real evaluated sub-solutions. After that, the real fitness improvement values of the sub-solutions therein are updated in steps 29-30 if a better overall solution is identified in steps 26-28. The steps 8-31 are repeated until meeting the termination condition which is generally set as the maximum number of allowed real evaluations.

From Algorithm 1, it can be seen that only $q/p \times 100\%$ percent of sub-solutions in RBF-SHADE-SACC need to be evaluated by the original simulation model, which significantly reduces the number of real evaluations since $q$ is generally much less than $p$. Besides, RBF-SHADE-SACC fits the SACC framework shown in Fig. 2 well. It can be utilized to verify the effectiveness of SACC on the one hand, and scales up RBF and SHADE to LSOPs on the other hand.

## 4 Experimental Studies

### 4.1 Experimental settings

The IEEE CEC 2010 benchmark suite [37] which contains 20 LSOPs was employed in our experiments. All these benchmark functions are minimization problems of 1000 dimensions. Table 1 presents their classification in terms of their separability, where the nonseparable sub-problems in partially separable functions all involve 50 decision variables. For more details about these functions, readers can refer to [37]. It is known that $F_{19}$ and $F_{20}$ are fully nonseparable functions and all kinds of CC algorithms show no advantage on them in comparison with traditional EAs, therefore they were excluded from our experiments.

In order to perform an unbiased analysis on RBF-SHADE-SACC, ideal decomposition was implemented, which means that all the decision variables of a benchmark function were manually grouped into some sub-problems according to the prior knowledge of the function. As suggested by [37], most existing CC algorithms take a maximum number of $3.0 \times 10^6$ FEs as the termination condition of a run. To show the superiority of RBF-SHADE-SACC, each of our experiments only employed 10 percent of the suggested computation resource, i.e., a maximum number of $3.0 \times 10^5$ FEs, as the default termination condition of a run. Unless otherwise mentioned, the result of each algorithm on a function was calculated based on 25 independent runs.

**Table 1** Classification of CEC 2010 benchmark functions

| Functions | Separability | No. of separable variables | No. of nonseparable variables |
|---|---|---|---|
| $F_1$ - $F_3$ | Separable | 1000 | $0 \times 50$ |
| $F_4$ - $F_8$ | Partially separable | 950 | $1 \times 50$ |
| $F_9$ - $F_{13}$ | Partially separable | 500 | $10 \times 50$ |
| $F_{14}$ - $F_{18}$ | Partially separable | 0 | $20 \times 50$ |
| $F_{19}$ - $F_{20}$ | Fully nonseparable | 0 | $1 \times 1000$ |

## 4.2 Parameter settings

There are several parameters in RBF-SHADE-SACC. Besides the archive size $d$ of RBF, the population size $p$ of $t$SHADE, and the number of elitist sub-solutions selected at each generation ($q$), it also needs another parameter $s$ to specify how to divide separable variables concerned in separable and partially separable functions such as $F_1$-$F_{13}$. It is understandable that separable variables could be divided in any way without affecting their theoretically optimal values. However, if all the separable variables are grouped into a single large scale sub-problem, then the advantage of CC will weaken and it will also raise the difficulty in building an accurate enough RBF model. On the contrary, if all the separable variables are independently treated, the difficulty in building RBF models will be alleviated, but the limited number of real FEs will have to be assigned to so many sub-problems that RBF-SHADE-SACC will be much likely not to converge on some sub-problems. From above analysis, it can be known that the key of dividing separable variables is to balance the difficulty in constructing RBF models and the quantity of computation resources assigned to each sub-problem. To numerically investigate the influence of $s$, we tested RBF-SHADE-SACC with different $s$ values selecting from {10, 20, 50, 100, 200}.

As for the archive size $d$ of RBF, its influence has been extensively analyzed at the end of section 3.2. Since RBF-SHADE-SACC continually updates the RBF model constructed for each sub-problem in an iteration-wise way, its sensitivity to $d$ could be greatly weakened. According to the suggestion given in [32], we set $d = 5s$ for the RBF related to each sub-problem. As a common parameter, the population size $p$ has been investigated much in the original SHADE [14], and it is revealed that the algorithm performs well on most small and medium scale problems when the population size is set to 100. This conclusion was also verified by our pilot experiments. Accordingly, $p$ was fixed to 100 for different sub-problems concerned in RBF-SHADE-SACC. As for $q$, it is a new parameter introduced by RBF-SHADE-SACC. The larger $q$ is, the more easily an accurate RBF model can be obtained since the more real evaluated sub-solutions will be employed to update the RBF archive at each generation, but the faster the available computation resource will be exhausted, which may be adverse to the convergence of RBF-SHADE-SACC. To numerically investigate the influence of $q$, we tested RBF-SHADE-SACC with different $q$ values selecting from {1, 5, 10, 15, 20}.

Taking the separable functions $F_1$ and $F_3$ and partially separable functions $F_{10}$ and $F_{13}$ as examples, Fig. 3 shows the average FVs obtained by RBF-SHADE-SACC when $s$ varies and $q$ is fixed to 10. It can be observed that, with the variation of $s$, RBF-SHADE-SACC shows sensitive and robust performances on the two separable functions $F_1$ and $F_3$, respectively. On the whole, a relatively small $s$ helps RBF-SHADE-SACC find superior solutions. However, an extremely small $s$ may destroy its performance on $F_1$. Due to this reason, we recommend to set $s = 20$ for separable functions. For each of the partially separable functions $F_{10}$ and $F_{13}$, RBF-SHADE-SACC needs to group the 500 separable variables therein according to $s$. Although its performance on these two functions is not affected too much by $s$ as that on $F_1$, it demonstrates completely opposite performance trends on these two functions when $s$ varies. Based on this observation, we recommend to set $s = 100$ for partially separable functions to balance the performance of RBF-SHADE-SACC on this kind of functions.

By keeping $s$ at the recommended value, Fig. 4 presents the influence of $q$. It is surprising to see that, for each of the test functions, the performance of RBF-SHADE-SACC generally improves with the increase of $q$ if it deteriorates with the increase of $s$, and vice verse. By comparing the basic functions concerned in these test functions, it can be revealed that the settings of $q$ and $s$ on LSOPs are directly affected by the landscape complexity of the basic functions. For example, the separable function $F_3$ takes Ackley function as its basic function which itself is a complicated multimodal function [37]. To generate a RBF model of acceptable accuracy for each sub-problem, RBF-SHADE-SACC has to reduce the dimension of each sub-problem and really evaluate more sub-solutions at each generation. On the contrary, the partially separable function $F_{13}$ takes Sphere function and Rosenbrock function as the basic functions for the separable and the nonseparable variables, respectively [37]. The landscapes of both basic functions are so simple that RBF-SHADE-SACC tends to increase $s$ and decrease $q$, so that it can undergo more generations without significantly sacrificing the accuracy of each RBF model. Despite this meaningful conclusion, we do not know the basic functions concerned in a black-box LSOP, not to mention their landscapes. For this reason, this study sets $q$ to a fixed value of 10 to balance the performance of RBF-SHADE-SACC on different functions.

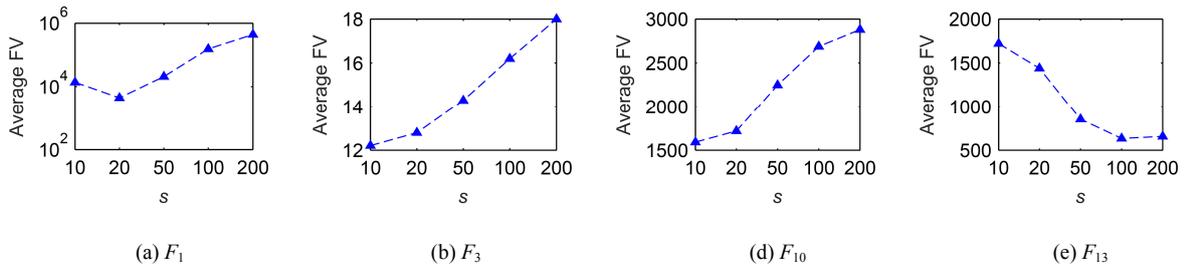

(a) $F_1$    (b) $F_3$    (d) $F_{10}$    (e) $F_{13}$

Fig. 3 Performance of RBF-SHADE-SACC with different $s$ values on $F_1$, $F_3$, $F_{10}$, and $F_{13}$

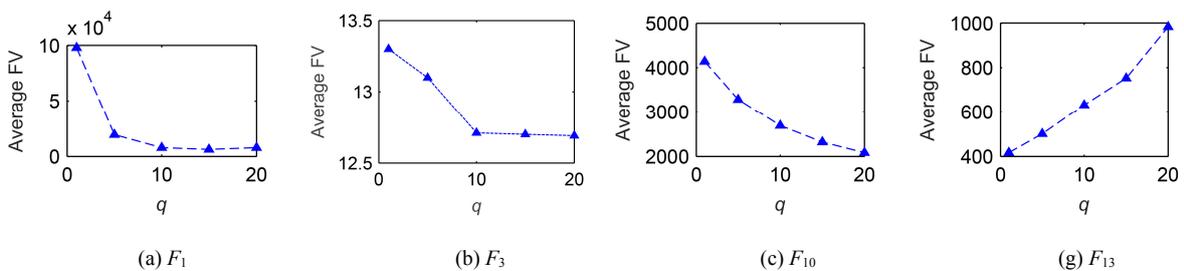

(a) $F_1$    (b) $F_3$    (c) $F_{10}$    (g) $F_{13}$

Fig. 4 Performance of RBF-SHADE-SACC with different $q$ values on $F_1$, $F_3$, $F_{10}$, and $F_{13}$

## 4.3 Comparison between RBF-SHADE-SACC and other CC algorithms

To evaluate the performance of RBF-SHADE-SACC, we specifically implemented a CC algorithm which takes the traditional CC framework and employs SHADE as optimizer. For the convenience of description, we name it SHADE-CC. The difference between RBF-SHADE-SACC and SHADE-CC mainly lies in the sub-solution evaluation method. Moreover, we also compared RBF-SHADE-SACC with an existing CC algorithm developed in [35] with a name of CC-I. Different from SHADE-CC, CC-I uses another efficient DE variant named SaNSDE as optimizer. Here CC-I serves as a baseline for SHADE-CC and RBF-SHADE-SACC. To ensure the fairness of the comparison, the parameters in SHADE-CC were set to the same values as the corresponding ones in RBF-SHADE-SACC. Table 2 summarizes the results obtained by SHADE-CC and RBF-SHADE-SACC with $1.0\times10^5$ and $3.0\times10^5$ FEs and the results obtained by CC-I with $3.0\times10^6$ FEs. It is necessary to mention that the results of CC-I are directly taken from [35]. To statistically analyze the performance of the three competitors, we employed Cohen's $d$ effect size [38] to quantify the difference among the average FVs obtained by them. Cohen's $d$ effect size is independent of the sample size and is generally considered 'small', 'medium', and 'large' if its absolute value belongs to [0.2, 0.3), [0.3, 0.8), and [0.8, +∞), respectively. According to this rule, if a result in Table 2 is judged to be better than, worse than, or similar to the corresponding one obtained by RBF-SHADE-SACC, it is marked with '+', '−', and '≈', respectively.

It can be seen from Table 2 that RBF-SHADE-SACC achieves excellent performance. When a maximum number of $3.0\times10^5$ FEs is allowed, it outperforms SHADE-CC on all of the 18 benchmark functions expect $F_6$ and $F_{11}$. Especially on $F_1$, $F_4$, $F_7$, $F_{12}$, and $F_{14}$, it yields better solutions than SHADE-CC by two orders of magnitude in terms of the average FV. As for $F_6$ and $F_{11}$, they take Ackley function as basic functions, whose fitness landscape is nearly a plateau in the solution region close to the global optimum and optimum is located in a very narrow region near the origin [37]. Then it is very difficult to build accurate enough RBF models for the sub-problems in $F_6$ and $F_{11}$ using a limited number of samples, which restricts the performance of RBF-SHADE-SACC on them. Even so, the best and median solutions obtained by RBF-SHADE-SACC on $F_6$ over 25 independent runs are much better than the corresponding ones obtained by SHADE-CC. When compared with CC-I, both RBF-SHADE-SACC and SHADE-CC show great superiority. They outperform CC-I on 15 and 12 out of total 18 functions. From the comparison between SHADE-CC and CC-I, it can be concluded that SHADE has an edge over SaNSDE for LSOPs under the CC framework, since the difference between SHADE-CC and CC-I mainly lies in optimizer and the former only consumes 10% of real FEs consumed by the latter. On the other side, the superiority of RBF-SHADE-SACC over SHADE-CC reveals that the RBF based sub-solution evaluation method is feasible and efficient.

When the maximum number of real FEs is reduced to $1.0\times10^5$, RBF-SHADE-SACC demonstrates more obvious advantage. It completely surpasses SHADE-CC and outperforms CC-I on all the benchmark functions except $F_{14}$, $F_{17}$, and

$F_{18}$. This result is very exciting since RBF-SHADE-SACC only consumes a thirtieth of real FEs consumed by CC-I. It also indicates that RBF-SHADE-SACC is more robust to the decreased computation resource than SHADE-CC.

Table 2 The results obtained by SHADE-CC, RBF-SHADE-SACC, and CC-I on CEC 2010 benchmark suite

| Fun. | No. of FEs | SHADE-CC | | | | | RBF-SHADE-CC | | | | | CC-I | |
|---|---|---|---|---|---|---|---|---|---|---|---|---|---|
| | | Best | Median | Worst | Mean | Std | Best | Median | Worst | Mean | Std | Mean | Std |
| $F_1$ | $1.0 \times 10^5$ | 1.43e+09 | 1.68e+09 | 1.85e+09 | 1.65e+09⁻ | 1.12e+08 | 5.32e+06 | 6.76e+06 | 9.29e+06 | 6.89e+06 | 1.03e+06 | 3.5e+11⁻ | 2.0e+10 |
| | $3.0 \times 10^5$ | 1.21e+07 | 1.32e+07 | 1.50e+07 | 1.34e+07⁻ | 7.70e+05 | 1.69e+03 | 2.89e+03 | 1.95e+04 | 4.33e+03 | 3.88e+03 | 3.5e+11⁻ | 2.0e+10 |
| $F_2$ | $1.0 \times 10^5$ | 6.91e+03 | 7.09e+03 | 7.31e+03 | 7.10e+03⁻ | 9.90e+01 | 1.58e+03 | 1.76e+03 | 2.13e+03 | 1.81e+03 | 1.46e+02 | 9.4e+03⁻ | 2.1e+02 |
| | $3.0 \times 10^5$ | 4.73e+03 | 4.86e+03 | 4.94e+03 | 4.84e+03⁻ | 6.17e+01 | 1.19e+03 | 1.29e+03 | 1.39e+03 | 1.29e+03 | 5.40e+01 | 9.4e+03⁻ | 2.1e+02 |
| $F_3$ | $1.0 \times 10^5$ | 1.68e+01 | 1.71e+01 | 1.73e+01 | 1.70e+01⁻ | 1.46e-01 | 1.38e+01 | 1.42e+01 | 1.46e+01 | 1.42e+01 | 1.98e-01 | 2.0e+01⁻ | 4.4e-02 |
| | $3.0 \times 10^5$ | 1.45e+01 | 1.48e+01 | 1.51e+01 | 1.48e+01⁻ | 1.66e-01 | 1.24e+01 | 1.28e+01 | 1.32e+01 | 1.28e+01 | 1.83e-01 | 2.0e+01⁻ | 4.4e−02 |
| $F_4$ | $1.0 \times 10^5$ | 1.23e+14 | 1.82e+14 | 2.18e+14 | 1.79e+14⁻ | 2.37e+13 | 1.96e+12 | 4.71e+12 | 9.31e+12 | 5.09e+12 | 1.77e+12 | 3.4e+14⁻ | 7.5e+13 |
| | $3.0 \times 10^5$ | 9.00e+12 | 2.09e+13 | 3.51e+13 | 2.13e+13⁻ | 5.86e+12 | 2.84e+11 | 7.44e+11 | 1.26e+12 | 7.42e+11 | 2.47e+11 | 3.4e+14⁻ | 7.5e+13 |
| $F_5$ | $1.0 \times 10^5$ | 3.66e+08 | 4.36e+08 | 4.71e+08 | 4.31e+08⁻ | 2.29e+07 | 7.16e+07 | 1.12e+08 | 1.53e+08 | 1.13e+08 | 2.31e+07 | 4.9e+08⁻ | 2.4e+07 |
| | $3.0 \times 10^5$ | 3.00e+08 | 3.35e+08 | 3.69e+08 | 3.34e+08⁻ | 1.81e+07 | 7.16e+07 | 1.12e+08 | 1.53e+08 | 1.13e+08 | 2.31e+07 | 4.9e+08⁻ | 2.4e+07 |
| $F_6$ | $1.0 \times 10^5$ | 4.21e+06 | 4.60e+06 | 5.12e+06 | 4.58e+06⁻ | 2.39e+05 | 1.66e+01 | 1.71e+01 | 1.56e+06 | 4.14e+05 | 6.35e+05 | 1.1e+07⁻ | 7.5e+05 |
| | $3.0 \times 10^5$ | 1.68e+03 | 3.02e+03 | 4.10e+03 | 3.04e+03⁺ | 6.22e+02 | 1.56e+01 | 1.61e+01 | 1.56e+06 | 4.14e+05 | 6.35e+05 | 1.1e+07⁻ | 7.5e+05 |
| $F_7$ | $1.0 \times 10^5$ | 3.07e+10 | 4.47e+10 | 5.97e+10 | 4.41e+10⁻ | 7.21e+09 | 2.70e+09 | 5.88e+09 | 9.76e+09 | 5.73e+09 | 1.92e+09 | 7.7e+10⁻ | 9.6e+09 |
| | $3.0 \times 10^5$ | 3.75e+09 | 7.52e+09 | 1.36e+10 | 7.67e+09⁻ | 2.52e+09 | 1.25e+06 | 1.03e+07 | 4.80e+08 | 6.23e+07 | 1.17e+08 | 7.7e+10⁻ | 9.6e+09 |
| $F_8$ | $1.0 \times 10^5$ | 4.64e+11 | 1.38e+12 | 4.01e+12 | 1.43e+12⁻ | 7.51e+11 | 3.57e+07 | 4.10e+07 | 2.63e+08 | 6.78e+07 | 5.64e+07 | 1.8e+14⁻ | 9.3e+13 |
| | $3.0 \times 10^5$ | 3.98e+07 | 5.11e+07 | 2.05e+08 | 7.61e+07⁻ | 4.38e+07 | 8.30e+06 | 1.79e+07 | 2.37e+08 | 3.99e+07 | 5.57e+07 | 1.8e+14⁻ | 9.3e+13 |
| $F_9$ | $1.0 \times 10^5$ | 2.68e+09 | 2.89e+09 | 3.17e+09 | 2.90e+09⁻ | 1.33e+08 | 7.25e+07 | 9.10e+07 | 1.04e+08 | 9.02e+07 | 9.08e+06 | 9.4e+08⁻ | 7.1e+07 |
| | $3.0 \times 10^5$ | 2.74e+08 | 3.45e+08 | 3.85e+08 | 3.41e+08⁻ | 2.92e+07 | 9.64e+06 | 1.19e+07 | 1.55e+07 | 1.19e+07 | 1.48e+06 | 9.4e+08⁻ | 7.1e+07 |
| $F_{10}$ | $1.0 \times 10^5$ | 9.17e+03 | 9.49e+03 | 9.66e+03 | 9.46e+03⁻ | 1.15e+02 | 2.47e+03 | 2.70e+03 | 2.90e+03 | 2.69e+03 | 1.28e+02 | 4.8e+03⁻ | 6.7e+01 |
| | $3.0 \times 10^5$ | 7.67e+03 | 7.82e+03 | 8.00e+03 | 7.82e+03⁻ | 9.34e+01 | 2.47e+03 | 2.70e+03 | 2.90e+03 | 2.69e+03 | 1.28e+02 | 4.8e+03⁻ | 6.7e+01 |
| $F_{11}$ | $1.0 \times 10^5$ | 9.67e+01 | 9.93e+01 | 1.03e+02 | 9.94e+01⁻ | 1.71e+00 | 1.98e+01 | 2.43e+01 | 2.76e+01 | 2.41e+01 | 1.97e+00 | 4.1e+01⁻ | 1.5e+00 |
| | $3.0 \times 10^5$ | 1.61e+01 | 1.66e+01 | 1.70e+01 | 1.66e+01⁺ | 2.43e-01 | 1.87e+01 | 2.31e+01 | 2.60e+01 | 2.28e+01 | 1.92e+00 | 4.1e+01⁻ | 1.5e+00 |
| $F_{12}$ | $1.0 \times 10^5$ | 7.15e+05 | 7.77e+05 | 8.10e+05 | 7.76e+05⁻ | 2.47e+04 | 1.13e+05 | 1.42e+05 | 1.77e+05 | 1.45e+05 | 1.80e+04 | 4.9e+05⁻ | 3.4e+04 |
| | $3.0 \times 10^5$ | 1.65e+05 | 1.86e+05 | 2.02e+05 | 1.85e+05⁻ | 1.14e+04 | 6.29e+02 | 1.51e+03 | 6.43e+03 | 1.89e+03 | 1.43e+03 | 4.9e+05⁻ | 3.4e+04 |
| $F_{13}$ | $1.0 \times 10^5$ | 1.79e+08 | 2.21e+08 | 2.93e+08 | 2.30e+08⁻ | 3.14e+07 | 7.81e+02 | 1.54e+03 | 6.15e+03 | 1.84e+03 | 1.05e+03 | 1.5e+07⁻ | 4.1e+06 |
| | $3.0 \times 10^5$ | 2.55e+03 | 3.71e+03 | 1.33e+04 | 4.14e+03⁻ | 2.25e+03 | 2.69e+02 | 5.89e+02 | 1.55e+03 | 6.32e+02 | 3.05e+02 | 1.5e+07⁻ | 4.1e+06 |
| $F_{14}$ | $1.0 \times 10^5$ | 5.94e+09 | 6.72e+09 | 7.36e+09 | 6.70e+09⁻ | 3.52e+08 | 2.46e+08 | 2.80e+08 | 3.20e+08 | 2.80e+08 | 2.02e+07 | 2.7e+07⁺ | 2.1e+06 |
| | $3.0 \times 10^5$ | 1.24e+09 | 1.37e+09 | 1.47e+09 | 1.36e+09⁻ | 6.37e+07 | 3.08e+07 | 3.60e+07 | 4.21e+07 | 3.59e+07 | 2.79e+06 | 2.7e+07⁺ | 2.1e+06 |
| $F_{15}$ | $1.0 \times 10^5$ | 9.33e+03 | 9.46e+03 | 9.61e+03 | 9.46e+03⁻ | 7.77e+01 | 1.98e+03 | 2.19e+03 | 2.32e+03 | 2.18e+03 | 7.89e+01 | 4.0e+03⁻ | 1.6e+02 |
| | $3.0 \times 10^5$ | 7.86e+03 | 8.06e+03 | 8.18e+03 | 8.04e+03⁻ | 9.38e+01 | 1.98e+03 | 2.19e+03 | 2.32e+03 | 2.18e+03 | 7.89e+01 | 4.0e+03⁻ | 1.6e+02 |
| $F_{16}$ | $1.0 \times 10^5$ | 2.29e+02 | 2.36e+02 | 2.43e+02 | 2.35e+02⁻ | 3.14e+00 | 5.93e+00 | 1.17e+01 | 1.87e+01 | 1.18e+01 | 3.28e+00 | 2.0e+01⁻ | 4.0e+00 |
| | $3.0 \times 10^5$ | 2.14e+01 | 2.33e+01 | 2.59e+01 | 2.33e+01⁻ | 1.23e+00 | 5.92e+00 | 1.16e+01 | 1.87e+01 | 1.18e+01 | 3.28e+00 | 2.0e+01⁻ | 4.0e+00 |
| $F_{17}$ | $1.0 \times 10^5$ | 1.58e+06 | 1.64e+06 | 1.76e+06 | 1.65e+06⁻ | 4.23e+04 | 3.89e+05 | 4.64e+05 | 5.44e+05 | 4.66e+05 | 3.54e+04 | 2.2e+01⁺ | 3.7e+01 |
| | $3.0 \times 10^5$ | 5.19e+05 | 5.58e+05 | 5.81e+05 | 5.56e+05⁻ | 1.68e+04 | 8.20e+03 | 1.29e+04 | 2.35e+04 | 1.38e+04 | 4.30e+03 | 2.2e+01⁺ | 3.7e+01 |
| $F_{18}$ | $1.0 \times 10^5$ | 2.86e+09 | 3.41e+09 | 4.10e+09 | 3.40e+09⁻ | 2.96e+08 | 3.27e+03 | 1.51e+04 | 3.21e+04 | 1.45e+04 | 6.39e+03 | 1.0e+03⁺ | 1.7e+02 |
| | $3.0 \times 10^5$ | 5.97e+04 | 9.51e+04 | 1.35e+05 | 9.31e+04⁻ | 1.87e+04 | 7.72e+02 | 1.37e+03 | 2.50e+03 | 1.43e+03 | 4.37e+02 | 1.0e+03⁺ | 1.7e+02 |
| No. of +/≈/− | $1.0 \times 10^5$ | | | 0/0/18 | | | | | − | | | 3/0/15 | |
| | $3.0 \times 10^5$ | | | 2/0/16 | | | | | − | | | 3/0/15 | |

To further verify the efficiency of RBF-SHADE-SACC, we conducted another experiment which counts the average number of FEs required by SHADE-CC to obtain the same average FV with RBF-SHADE-SACC. Table 3 presents the results when RBF-SHADE-SACC is allowed to undergo $1.0 \times 10^5$ and $3.0 \times 10^5$ real FEs. It can be seen from Table 3 that, to achieve similar results, RBF-SHADE-SACC generally requires much fewer real FEs than SHADE-CC. When RBF-SHADE-SACC is allowed to undergo $1.0 \times 10^5$ real FEs, SHADE-CC demands three times of computation resource at least to achieve similar results on all the benchmark functions except $F_6$ and $F_{11}$. Especially on $F_2$, $F_5$, $F_{10}$, and $F_{15}$, it consumes more than ten times of computation resource. When a maximum number of $3.0 \times 10^5$ real FEs is taken as the termination condition for RBF-SHADE-SACC, the inferiority of SHADE-CC is alleviated to a certain extent, but it still requires at least three times of computation resource on 10 out of the total 18 benchmark functions and at least two times of computation resource on 5 out of the other 8 benchmark functions.

**Table 3** The number of FEs required by SHADE-CC to achieve the same result with RBF-SHADE-SACC

| No. of real FEs | $F_1$ | $F_2$ | $F_3$ | $F_4$ | $F_5$ | $F_6$ | $F_7$ | $F_8$ | $F_9$ |
|---|---|---|---|---|---|---|---|---|---|
| $1.0 \times 10^5$ FEs | $3.32 \times 10^5$ | $1.37 \times 10^6$ | $4.38 \times 10^5$ | $5.57 \times 10^5$ | $1.39 \times 10^6$ | $1.83 \times 10^5$ | $3.30 \times 10^5$ | $4.11 \times 10^5$ | $5.75 \times 10^5$ |
| $3.0 \times 10^5$ FEs | $6.94 \times 10^5$ | $1.62 \times 10^6$ | $1.07 \times 10^6$ | $1.14 \times 10^6$ | $1.39 \times 10^6$ | $1.83 \times 10^5$ | $7.23 \times 10^5$ | $7.14 \times 10^5$ | $1.23 \times 10^6$ |
| No. of real FEs | $F_{10}$ | $F_{11}$ | $F_{12}$ | $F_{13}$ | $F_{14}$ | $F_{15}$ | $F_{16}$ | $F_{17}$ | $F_{18}$ |
| $1.0 \times 10^5$ FEs | $1.37 \times 10^6$ | $2.39 \times 10^5$ | $3.39 \times 10^5$ | $3.83 \times 10^5$ | $6.78 \times 10^5$ | $2.64 \times 10^6$ | $3.29 \times 10^5$ | $3.35 \times 10^5$ | $3.96 \times 10^5$ |
| $3.0 \times 10^5$ FEs | $1.37 \times 10^6$ | $2.39 \times 10^5$ | $9.46 \times 10^5$ | $7.96 \times 10^5$ | $1.47 \times 10^6$ | $2.68 \times 10^6$ | $3.29 \times 10^5$ | $1.03 \times 10^6$ | $8.96 \times 10^5$ |

In order to examine the evolution characteristics of RBF-SHADE-SACC, Fig. 5 compares the evolution curves of the average FVs obtained by RBF-SHADE-SACC and SHADE-CC, where functions $F_1$, $F_3$, $F_{10}$, and $F_{13}$ are taken as examples. From Fig. 5, two phenomena can be observed. On the one hand, RBF-SHADE-SACC performs slightly worse than SHADE-CC at the initial stage of the evolution process. The reason mainly consists in that RBF-SHADE-SACC generates and really evaluates much more random sub-solutions to construct RBF model for each sub-problem at its initial search stage, while SHADE-CC quickly moves to the optimization stage after the simple initialization of each sub-population. On the other hand, RBF-SHADE-SACC yields better solutions than SHADE-CC after several hundreds of real FEs and keeps a better evolution trend until all the available computation resources are exhausted.

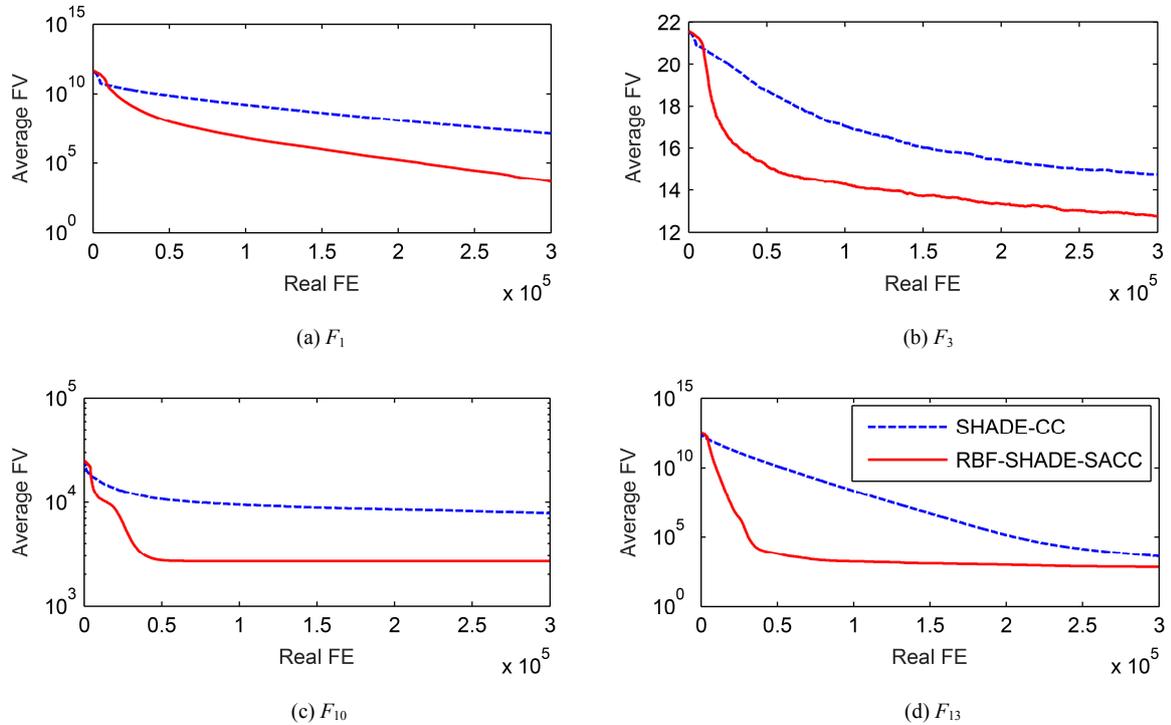

**Fig. 5** The evolution trends of average FVs obtained by RBF-SHADE-SACC and SHADE-CC on $F_1$, $F_3$, $F_{10}$ and $F_{13}$

To sum up, RBF-SHADE-SACC shows better performance than SHADE-CC and CC-I on most test functions. The success of RBF-SHADE-SACC benefits from three algorithmic components, including the SACC framework, the RBF model, and the SHADE algorithm. The SACC framework provides basic design rules for RBF-SHADE-SACC by introducing surrogate models into the traditional CC, and enables RBF-SHADE-SACC to generate much more candidate solutions when a limited number of computation resources are available. Consequently, the solution space can be explored more thoroughly. The RBF model is easy to train and becomes more and more accurate with the update of the model archive, and thus can efficiently evaluate a huge number of sub-solutions and filter out the inferior sub-solutions. After some simple modifications, SHADE cooperates well with RBF model under the SACC framework and can efficiently solve the lower dimensional sub-problems obtained through decomposition.

## 5 Conclusion

In this paper, a novel CC framework named SACC is proposed for LSOPs. SACC is mainly characterized by employing a surrogate model based sub-solution evaluation method. Different from the traditional CC which evaluates each sub-solution based on a context vector and the original time-consuming simulation model, SACC builds a calculable surrogate model for each sub-problem and employs it to filter out most of the inferior sub-solutions. It does not invoke the original simulation model except evaluating some high quality sub-solutions. As a result, its requirement on computation resource can be greatly reduced. This paper also designs a concrete SACC algorithm by introducing RBF and SHADE into the SACC framework. To make RBF and SHADE adapt to the characteristics of LSOP and SACC, some modifications are conducted on them. Experimental results on CEC 2010 benchmark suite demonstrate that RBF significantly improves the sub-solution evaluation efficiency, and compared with the traditional CC algorithms, the resultant RBF-SHADE-SACC can generate highly competitive solutions even with much fewer computation resources.

The SACC algorithm presented in this paper employs the same surrogate model and the same optimizer for all the sub-problems which may have strikingly different characteristics. Our future work will focus on developing SACC algorithms which can adaptively configure surrogate models and optimizers according to the characteristics of sub-problems. Moreover, we will further verify the efficiency of SACC on other benchmark functions and some real world problems.